\documentclass[conference,10pt]{IEEEtran}

\usepackage{cite}
\usepackage{amsmath,amssymb,amsfonts}
\usepackage{graphicx}
\usepackage{textcomp}
\usepackage{xcolor}
\usepackage{booktabs}
\usepackage{tabularx}
\usepackage{url}

\def\BibTeX{{\rm B\kern-.05em{\sc i\kern-.025em b}\kern-.08em
    T\kern-.1667em\lower.7ex\hbox{E}\kern-.125emX}}

\begin{document}

\title{Beyond Baseline Severity: Temporal and Disease-Specific Predictors of Depression Outcomes Following Mindfulness Interventions}

\author{
\IEEEauthorblockN{
Muhammad Jawad Chowdhury\IEEEauthorrefmark{1},
Sultanus Salehin\IEEEauthorrefmark{2},
Akib Jayed Islam\IEEEauthorrefmark{3}
}
\IEEEauthorblockA{\IEEEauthorrefmark{1}Department of Computer Science \& Engineering, Islamic University of Technology, Bangladesh}
\IEEEauthorblockA{\IEEEauthorrefmark{2}Department of Electrical \& Electronic Engineering, Islamic University of Technology, Bangladesh}
\IEEEauthorblockA{\IEEEauthorrefmark{3}Computational Colour \& Spectral Imaging, Norwegian University of Science and Technology, Norway}
\IEEEauthorblockA{
Email: \{jawad, sultanussalehin\}@iut-dhaka.edu, akibjayedislam@gmail.com
}
}

\maketitle

\begin{abstract}
Depression severity among patients with chronic or acute medical conditions is influenced by a complex interaction of baseline psychological state, demographic characteristics, clinical context, and engagement with behavioral interventions. This paper presents an interpretable machine-learning analysis of a multi-center longitudinal clinical cohort to predict Beck Depression Inventory-II (BDI-II) scores at 12 and 24 weeks following mindfulness-based intervention participation. The study uses demographic variables, clinical condition information, hospital-center identifiers, baseline BDI-II scores, and therapy engagement measures to model short-term and long-term depression outcomes. Missing follow-up outcomes were addressed using a model-based stochastic imputation procedure to preserve the modest sample size while maintaining outcome variability. Five regression models were evaluated, spanning regularized linear regression and tree-based ensemble methods. Ridge Regression achieved the best 12-week performance with an RMSE of 5.186 and $R^2$ of 0.474, while LightGBM achieved the best 24-week performance with an RMSE of 5.038 and $R^2$ of 0.525. Beyond prediction accuracy, the analysis reveals three clinically relevant patterns: baseline severity remains the strongest overall predictor, short-term outcomes are more strongly associated with clinical and hospital context, and long-term outcomes show greater dependence on behavioral adherence and demographic factors. Disease-specific and hierarchical subgroup analyses further indicate that predictors differ substantially across and within clinical categories. These findings support the use of interpretable, context-aware modeling to inform personalized mental-health support following mindfulness-based interventions.
\end{abstract}


\section{Introduction}

\subsection{Clinical Motivation}
Depression is a major contributor to global disability and frequently co-occurs with chronic and acute physical illness, with comorbidity associated with poorer health status, reduced treatment adherence, and lower quality of life \cite{vos2020global,katon2011epidemiology}. In clinical populations, depressive symptoms are not static: they may improve, persist, relapse, or recur depending on baseline psychological burden, disease context, demographic vulnerability, and engagement with treatment or supportive interventions \cite{fournier2010antidepressant,benrimoh2023outcome}. Consequently, longitudinal depression assessment is important not only for identifying patients with elevated symptom severity, but also for monitoring treatment response, detecting residual symptoms, and understanding how risk evolves after intervention \cite{trivedi2006evaluation,fortney2017tipping}. Clinical guidelines have emphasized the value of depression screening in adult populations, particularly when adequate systems exist for diagnosis, treatment, and follow-up \cite{siu2016screening,uspstf2023screening}.

The Beck Depression Inventory-II (BDI-II) is a widely used self-report instrument for measuring depressive symptom severity and monitoring changes over time \cite{beck1996bdi}. In longitudinal intervention studies, repeated BDI-II measurements provide a practical way to quantify depressive symptom severity, monitor change over time, and examine both short-term response and longer-term maintenance of benefit \cite{beck1996bdi,teasdale2000prevention}. However, predicting future BDI-II scores remains challenging because depression trajectories are heterogeneous and may be shaped by interacting clinical, demographic, behavioral, and care-context factors \cite{dwyer2018machine,shatte2019machine,benrimoh2023outcome}. For example, a patient with high baseline depression severity may remain at elevated risk despite intervention, while another patient with similar baseline severity may improve substantially due to better therapy engagement or more favorable clinical context. This motivates the use of predictive and interpretable modeling approaches that can identify not only whether future depression severity can be estimated, but also which factors drive those estimates.

\subsection{Prior Work}
Mindfulness-based interventions have been widely studied as non-pharmacological approaches for reducing depressive symptoms and preventing relapse \cite{hofmann2010effect,kuyken2016efficacy}. Mindfulness-based interventions integrate attentional training, present-moment awareness, and cognitive self-regulation practices to help patients recognize and disengage from maladaptive thought patterns \cite{segal2002mindfulness,teasdale2000prevention}. Early clinical evidence showed that Mindfulness-Based Cognitive Therapy can reduce relapse or recurrence among patients with major depression \cite{teasdale2000prevention}. Subsequent systematic reviews and meta-analyses have further supported the effectiveness of mindfulness-based approaches for relapse prevention, particularly among individuals with recurrent depression \cite{piet2011effect,kuyken2016efficacy}. These findings suggest that mindfulness participation may play an important role in depression management, but they also raise a clinically important question: are the effects and predictors of improvement uniform across patient groups?

For longitudinal depression assessment, machine-learning models offer a way to combine baseline symptom severity, demographic characteristics, clinical context, and intervention-engagement variables within a single predictive framework \cite{shatte2019machine,dwyer2018machine}. This is particularly useful when outcome trajectories are shaped by non-linear relationships or interactions among patient-level and care-context variables. However, clinical usefulness depends not only on predictive accuracy but also on interpretability, actionability, and fit with clinical decision-making workflows \cite{shah2019making,markus2021role}. Black-box predictions alone provide limited guidance for clinicians unless the model can support clinically meaningful explanation, indicate influential patient-level factors, and help users understand why a prediction was produced \cite{rudin2019stop,caruana2015intelligible}. Explainable machine-learning methods, including coefficient-based interpretation for regularized linear models and SHAP-based explanations for non-linear models, can help translate predictive models into clinically meaningful insights \cite{lundberg2017unified}.

\subsection{Research Gap and Contributions}
Despite growing evidence on mindfulness-based interventions and increasing use of machine learning in mental health, two gaps remain important. First, many predictive studies emphasize overall performance while paying less attention to how predictor importance changes over time. Short-term depression outcomes may be driven by immediate clinical context, whereas long-term outcomes may depend more strongly on sustained behavioral engagement and demographic vulnerability. Second, global models may obscure disease-specific heterogeneity. Patients with cancer, cardiovascular disease, renal insufficiency, or other clinical conditions may experience different psychological burdens and different barriers to intervention engagement. Therefore, a single global model may be insufficient for understanding depression trajectories across clinically diverse populations.

This paper addresses these gaps through an interpretable machine-learning analysis of a multi-center longitudinal clinical cohort. We predict BDI-II depression severity at 12 and 24 weeks using demographic variables, clinical condition information, hospital-center identifiers, baseline BDI-II scores, and mindfulness therapy engagement measures. The main contributions are threefold. First, we develop and evaluate multiple regression models, including Ridge Regression and tree-based ensemble methods, for short-term and long-term BDI-II prediction. Second, we compare feature importance across 12-week and 24-week outcomes to identify temporal shifts in the drivers of depression severity. Third, we perform disease-specific and hierarchical subgroup analyses to examine whether predictor profiles differ across and within clinical categories. Through this framework, the study reframes depression outcome prediction as an interpretable, context-aware modeling problem and provides evidence that personalized intervention strategies may be more appropriate than a single uniform approach.

\section{Dataset Overview and Exploratory Analysis}
\label{sec:dataset_eda}

\subsection{Dataset and Missingness}
The study uses a multi-center longitudinal clinical dataset containing 210 patient records. Depression severity was measured using the Beck Depression Inventory-II (BDI-II) at baseline, 12 weeks, and 24 weeks. The 12-week score represents short-term post-intervention severity, while the 24-week score represents longer-term follow-up severity. The predictors include demographic variables, clinical condition information, hospital-center identifiers, and mindfulness intervention engagement measures.

The dataset contains limited missingness in baseline predictors but more substantial missingness in follow-up outcomes. Specifically, \texttt{age}, \texttt{condition\_type}, and \texttt{bdi\_ii\_baseline} each contain one missing value, while 43 records are missing the 12-week BDI-II score and 44 records are missing the 24-week BDI-II score. Because listwise deletion would reduce the effective sample size and weaken subgroup analysis, missing outcome values were handled using the model-based stochastic imputation strategy described in Section~\ref{sec:methodology}.

\begin{table}[t]
\centering
\caption{Summary of key variables and missingness.}
\label{tab:dataset_summary}
\footnotesize
\begin{tabularx}{\columnwidth}{p{0.31\columnwidth} X c}
\toprule
\textbf{Category} & \textbf{Variables} & \textbf{Missing} \\
\midrule
Demographic & Age, sex & 1, 0 \\
Clinical & Condition, condition type & 0, 1 \\
Institutional & Hospital center identifier & 0 \\
Baseline outcome & Baseline BDI-II & 1 \\
Follow-up outcomes & 12-week BDI-II, 24-week BDI-II & 43, 44 \\
Intervention engagement & Therapies started, therapies completed & 0, 0 \\
\bottomrule
\end{tabularx}
\end{table}


\subsection{Exploratory Patterns}
\begin{figure}[t]
    \centering
    \includegraphics[width=0.95\columnwidth]{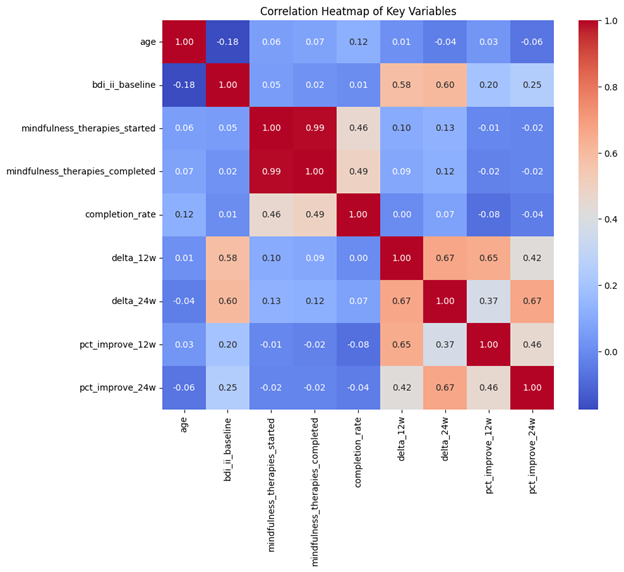}
    \caption{Correlation heatmap of key variables. Baseline BDI-II shows positive associations with both 12-week and 24-week outcomes, while therapy engagement variables show weaker linear associations with symptom change.}
    \label{fig:correlation_heatmap}
\end{figure}

As shown in Fig.~\ref{fig:correlation_heatmap}, exploratory analysis indicated that baseline depression severity was strongly associated with future BDI-II outcomes. Baseline BDI-II showed positive correlations with both 12-week BDI-II ($r=0.56$) and 24-week BDI-II ($r=0.48$), supporting initial symptom burden as an important predictor of later depression severity. However, these correlations also indicate that baseline severity alone does not fully explain follow-up outcomes.

Clinical subgroup analysis revealed clear heterogeneity across condition types. Follow-up BDI-II distributions differed across disease groups, suggesting that depression trajectories after mindfulness intervention may depend on the patient's underlying clinical context. In the observed data, renal-related subgroups generally showed lower follow-up depression scores, whereas the Breast Cancer subgroup showed higher median follow-up scores. Because some subtypes have limited sample sizes, these patterns are interpreted as exploratory rather than definitive.

\begin{figure}[t]
    \centering
    \includegraphics[width=\columnwidth]{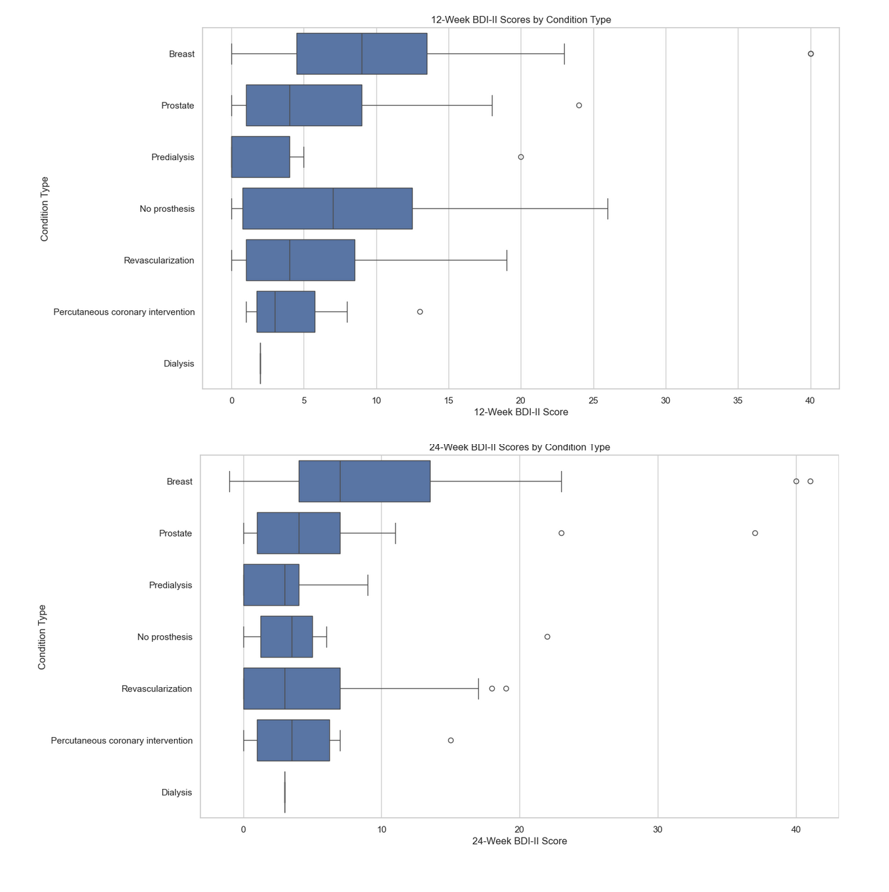}
    \caption{BDI-II scores at 12 weeks and 24 weeks stratified by condition type. Variation across condition types suggests disease-specific differences in depression outcomes.}
    \label{fig:condition_box}
\end{figure}

Therapy engagement showed a more complex pattern than a simple global linear association with improvement. Although higher completion may be beneficial for some patients, the relationship between completion rate and symptom change was not uniform across the full cohort. This suggests that adherence may interact with baseline severity, disease type, and hospital context. These exploratory findings motivate the subsequent predictive modeling, temporal comparison, and disease-specific interpretability analyses.

\section{Methodology}
\label{sec:methodology}

\subsection{Problem Formulation}
Let $\mathbf{x}_i \in \mathbb{R}^{p}$ denote the feature vector for patient $i$, including demographic, clinical, hospital-center, baseline-severity, and therapy-engagement variables. The prediction targets are the BDI-II scores at 12 weeks and 24 weeks, denoted by $y_i^{(12)}$ and $y_i^{(24)}$, respectively. For each prediction horizon $t \in \{12,24\}$, the task is formulated as a supervised regression problem:
\begin{equation}
    \hat{y}_i^{(t)} = f_t(\mathbf{x}_i),
\end{equation}
where $\hat{y}_i^{(t)}$ is the predicted BDI-II score and $f_t(\cdot)$ is the learned prediction function for horizon $t$.

Model performance was evaluated on the held-out test set using Root Mean Squared Error (RMSE) and the coefficient of determination ($R^2$). RMSE was used to quantify prediction error in BDI-II score units, while $R^2$ was used to assess the proportion of outcome variance explained by each model.

\subsection{Preprocessing and Feature Engineering}
Missing values in baseline predictors were handled using simple imputation: missing continuous values were replaced with the median and missing categorical values with the mode. Categorical variables, including sex, condition, condition type, and hospital-center identifier, were transformed using one-hot encoding. Continuous variables were standardized to zero mean and unit variance for models sensitive to feature scaling.

A therapy-adherence feature was engineered to capture the fraction of mindfulness sessions completed:
\begin{equation}
    r_i = \frac{c_i}{s_i + \epsilon},
\end{equation}
where $c_i$ is the number of mindfulness therapies completed by patient $i$, $s_i$ is the number of therapies started, and $\epsilon$ is a small constant used to avoid division by zero. This feature was used to represent behavioral engagement with the intervention.

\subsection{Handling Missing Outcome Data}
A major methodological challenge was missingness in the 12-week and 24-week BDI-II outcomes. Since deleting all records with missing follow-up scores would reduce the already modest sample size, a model-based stochastic imputation strategy was adopted. For each outcome horizon, a Random Forest helper model was trained on records with observed target values and then used to estimate missing follow-up scores:
\begin{equation}
    \tilde{y}_i^{(t)} = g_t(\mathbf{x}_i), \quad t \in \{12,24\},
\end{equation}
where $g_t(\cdot)$ is the helper model for the corresponding outcome. To avoid artificially shrinking outcome variance, stochastic noise was applied to the imputed values:
\begin{equation}
    y_{i,\mathrm{imp}}^{(t)} = \tilde{y}_i^{(t)} \cdot \eta_i,
\end{equation}
where $\eta_i \sim U(0.8,1.2)$.

This approach preserved the full cohort for subsequent modeling while introducing variability into imputed outcomes. However, because the completed dataset was generated before final train-test evaluation, results should be interpreted as exploratory and hypothesis-generating rather than as a strictly leakage-isolated validation.

\subsection{Predictive Models}
Five regression models were evaluated: Ridge Regression, Random Forest, XGBoost, LightGBM, and CatBoost. Ridge Regression was included as an interpretable linear baseline with $L_2$ regularization:
\begin{equation}
    \min_{\boldsymbol{\beta}} \sum_{i=1}^{n}(y_i-\boldsymbol{\beta}^{T}\mathbf{x}_i)^2 + \lambda \|\boldsymbol{\beta}\|_2^2,
\end{equation}
where $\lambda$ controls the strength of regularization. Tree-based ensemble models were included to capture non-linear relationships and interactions among clinical, demographic, and therapy-engagement variables.

The completed dataset was split into 80\% training and 20\% testing subsets using a fixed random seed. The same evaluation metrics were applied for both 12-week and 24-week prediction tasks.

\subsection{Interpretability and Subgroup Analysis}
Interpretability was examined at three levels. First, global feature importance was used to identify overall predictors across all patients. For Ridge Regression, standardized coefficients were used to determine feature direction and magnitude. For LightGBM, SHAP values were used to quantify feature contributions in the non-linear model. Second, temporal comparison was performed by contrasting important predictors for 12-week and 24-week outcomes. Third, disease-specific models were trained for selected clinical subgroups to examine whether predictors differed by condition type. A hierarchical subgroup analysis was also performed within broader disease categories, particularly cancer, to compare subtype-specific predictor profiles.

\section{Predictive Modeling Results}

\subsection{Global Model Performance}
Table~\ref{tab:final_performance} summarizes model performance for 12-week and 24-week BDI-II outcomes. The best model at each horizon was selected using the lowest RMSE, with $R^2$ used as a complementary measure of explained variance.

The regularized linear model achieved the strongest 12-week performance, with an RMSE of 5.186 and an $R^2$ of 0.474. In contrast, the non-linear gradient-boosting model achieved the strongest 24-week performance, with an RMSE of 5.038 and an $R^2$ of 0.525. This difference suggests that short-term outcomes may be captured reasonably well by additive relationships among baseline, clinical, and engagement-related variables, whereas longer-term outcomes may involve more complex non-linear interactions.

\begin{table}[t]
\centering
\caption{Final model performance on the imputed dataset for 12-week and 24-week BDI-II prediction.}
\label{tab:final_performance}
\begin{tabular}{lcccc}
\toprule
\textbf{Model} & \multicolumn{2}{c}{\textbf{12 Weeks}} & \multicolumn{2}{c}{\textbf{24 Weeks}} \\
 & \textbf{RMSE} & \textbf{$R^2$} & \textbf{RMSE} & \textbf{$R^2$} \\
\midrule
Ridge Regression & \textbf{5.186} & \textbf{0.474} & 6.044 & 0.317 \\
Random Forest    & 5.972 & 0.303 & 5.762 & 0.379 \\
XGBoost          & 6.053 & 0.284 & 6.031 & 0.320 \\
LightGBM         & 5.441 & 0.421 & \textbf{5.038} & \textbf{0.525} \\
CatBoost         & 6.006 & 0.295 & 5.398 & 0.455 \\
\bottomrule
\end{tabular}
\end{table}

\subsection{Temporal Impact Analysis}
The temporal comparison revealed a clear shift in predictor importance between short-term and long-term outcomes. For the 12-week regularized linear model, feature importance was interpreted using the absolute value of standardized model coefficients:
\begin{equation}
    I_{j}^{(12)} = \left|\beta_{j}^{(12)}\right|,
\end{equation}
where $\beta_{j}^{(12)}$ is the standardized coefficient of feature $j$ in the 12-week model. For the 24-week non-linear model, feature importance was summarized using mean absolute SHAP values:
\begin{equation}
    I_{j}^{(24)} = \frac{1}{n}\sum_{i=1}^{n}\left|\phi_{ij}^{(24)}\right|,
\end{equation}
where $\phi_{ij}^{(24)}$ denotes the SHAP contribution of feature $j$ for patient $i$ in the 24-week model.

Because coefficient magnitudes and SHAP values are measured on different explanation scales, temporal comparison was based on relative importance patterns rather than direct raw magnitude comparison. For each prediction horizon, importance scores can be normalized as
\begin{equation}
    \widetilde{I}_{j}^{(t)} =
    \frac{I_{j}^{(t)}}{\sum_{k=1}^{p} I_{k}^{(t)}} ,
\end{equation}
where $p$ is the number of predictors. The temporal change in relative importance for feature $j$ can then be represented as
\begin{equation}
    \Delta_j^{(24-12)} =
    \widetilde{I}_{j}^{(24)} - \widetilde{I}_{j}^{(12)} .
\end{equation}
Positive values of $\Delta_j^{(24-12)}$ indicate features that become relatively more influential at 24 weeks, whereas negative values indicate features that are more influential at 12 weeks.

For the 12-week model, clinical and contextual variables were highly influential. Hospital-center indicators and condition-specific variables appeared among the most important predictors, indicating that early depression outcomes may be strongly shaped by immediate care environment and disease context. Baseline BDI-II severity remained a major predictor, supporting that initial psychological burden strongly influences short-term symptom levels.

For the 24-week model, the strongest predictors shifted toward baseline state, age, and therapy-engagement variables. Session completion rate and the number of therapies started or completed became more prominent at 24 weeks than at 12 weeks. This suggests that long-term depression trajectories may depend more on sustained engagement and demographic vulnerability, whereas short-term outcomes are more strongly influenced by clinical and institutional context.

\begin{figure}[t]
    \centering
    \includegraphics[width=\columnwidth]{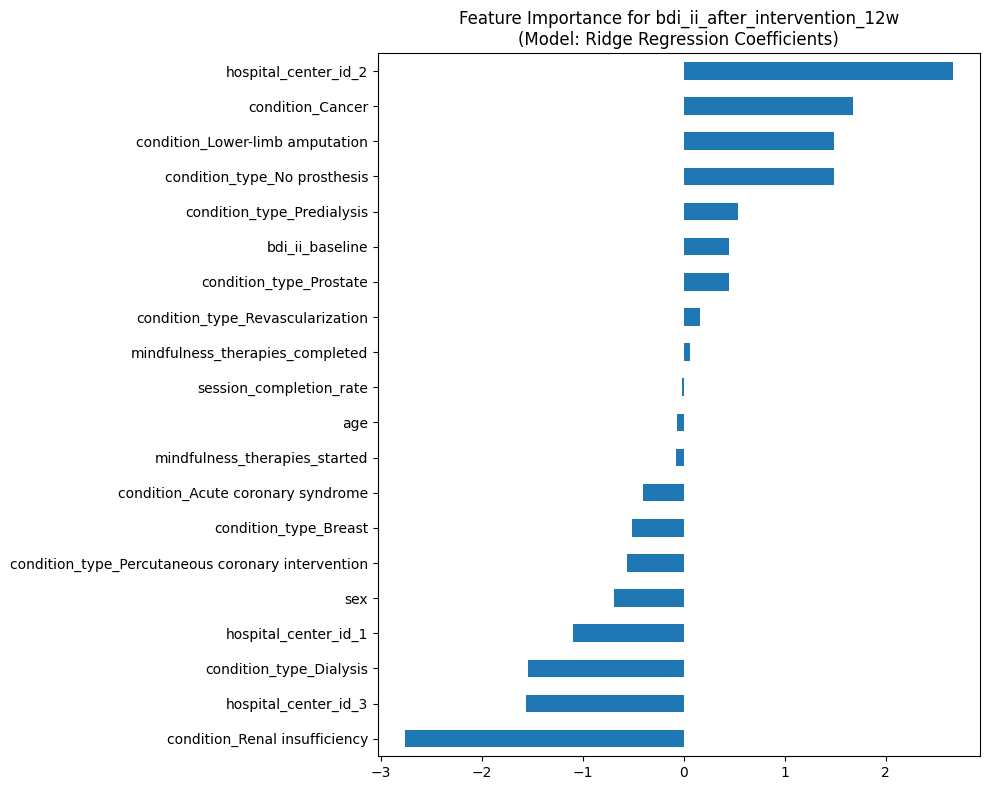}
    \caption{Feature importance for 12-week BDI-II prediction based on standardized Ridge Regression coefficients.}
    \label{fig:temporal_12w}
\end{figure}

\begin{figure}[t]
    \centering
    \includegraphics[width=\columnwidth]{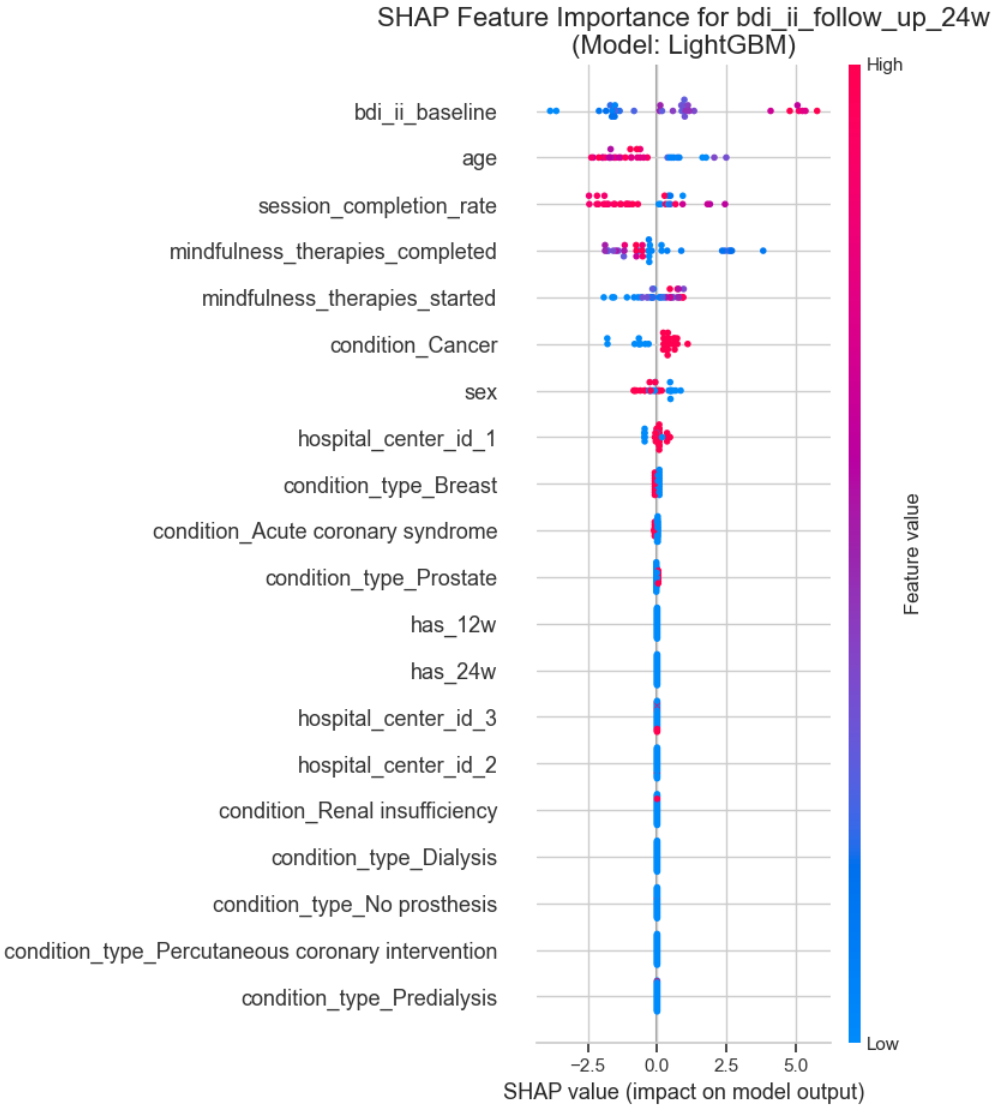}
    \caption{SHAP feature importance for 24-week BDI-II prediction based on the LightGBM model.}
    \label{fig:temporal_24w}
\end{figure}

Overall, the temporal analysis indicates that depression outcome predictors are not static. Initial clinical context appears more important for immediate post-intervention outcomes, while behavioral adherence and demographic factors become more influential in longer-term follow-up.

\subsection{Disease-Specific Comparison}
To examine whether global patterns masked subgroup-specific effects, dedicated models were trained for selected disease groups and interpreted separately within each clinical context. This allowed predictor importance to be compared across subgroups rather than assuming a single global relationship for all patients.

The results revealed substantial heterogeneity across clinical subgroups. For patients with Acute Coronary Syndrome, therapy adherence was one of the most important predictors of short-term improvement. In contrast, Breast Cancer outcomes were dominated by baseline BDI-II severity, indicating that initial psychological burden was a stronger determinant of follow-up scores in this subgroup. For Renal Insufficiency, hospital-center identifiers were among the most influential predictors, suggesting that institutional or care-process differences may be especially relevant for this population.

\begin{figure}[t]
    \centering
    \includegraphics[width=0.95\columnwidth]{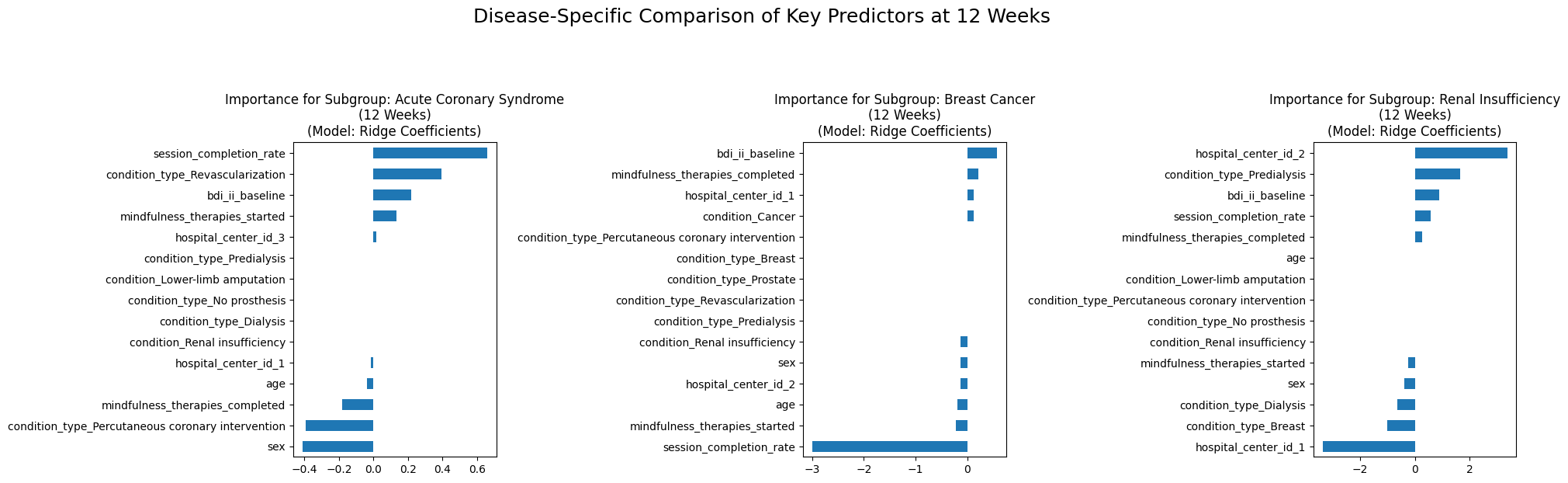}
    \caption{Disease-specific comparison of key predictors across three clinical subgroups at 12 weeks. Predictor patterns differ substantially by clinical context.}
    \label{fig:disease_comparison_12w}
\end{figure}

These findings show that a single global interpretation is insufficient. While baseline severity is consistently important, the relative influence of adherence, hospital context, and clinical condition varies across disease groups. This supports the need for subgroup-aware predictive modeling when interpreting mental-health outcomes in clinically heterogeneous populations.

\subsection{Hierarchical Subgroup Analysis}
A hierarchical analysis was conducted to examine whether heterogeneity persists within broader disease categories. Within the cancer cohort, separate subtype-level interpretations were compared for Breast Cancer and Prostate Cancer.

\begin{figure}[b]
    \centering
    \includegraphics[width=0.95\columnwidth]{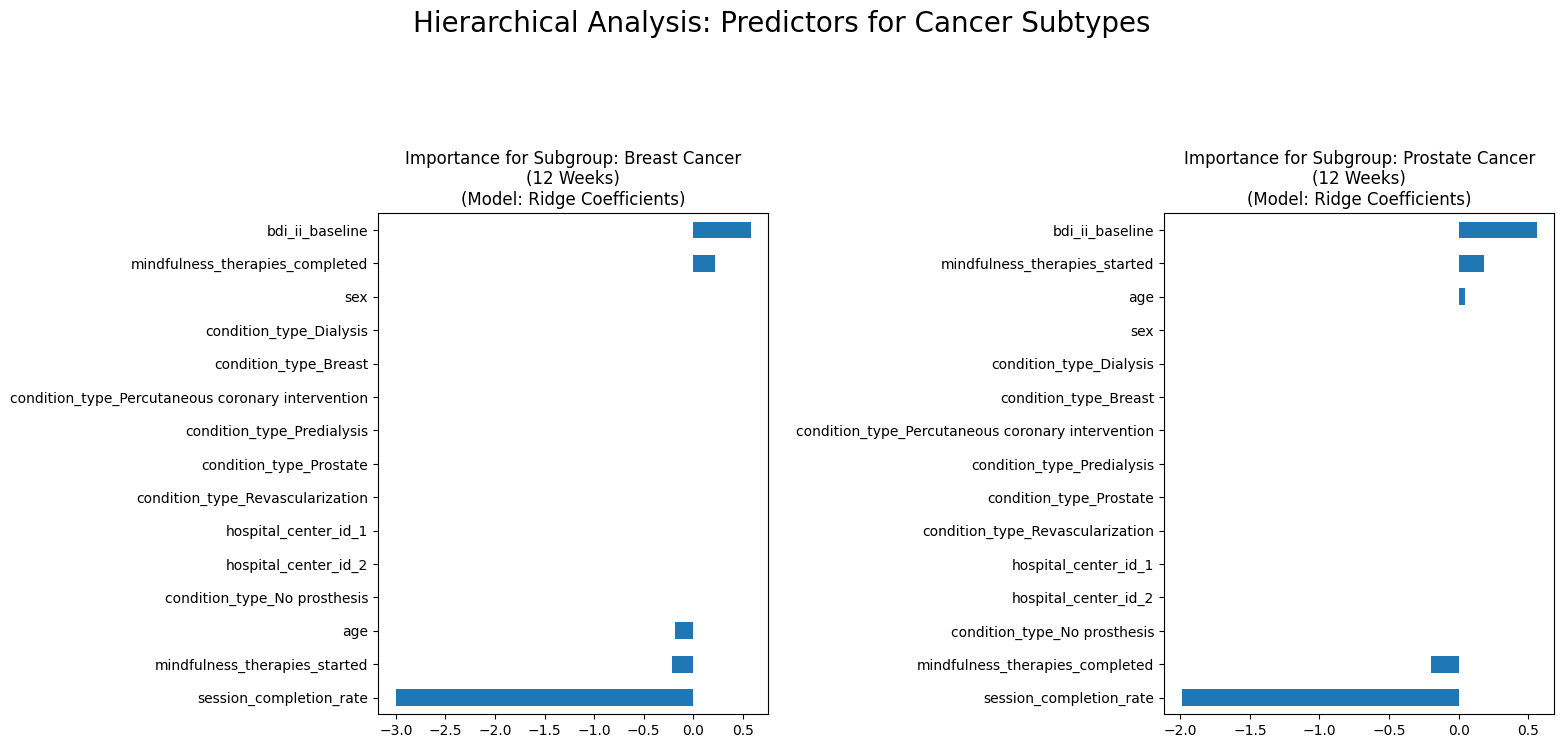}
    \caption{Hierarchical comparison of predictors for Breast Cancer and Prostate Cancer subtypes at 12 weeks. Heterogeneity persists within the broader cancer category.}
    \label{fig:cancer_hierarchy}
\end{figure}

The analysis showed distinct predictor profiles across these cancer subtypes. For both subgroups, baseline BDI-II severity contributed to worse predicted outcomes. However, behavioral engagement measures such as session completion rate and number of therapies started were relatively more important for Breast Cancer, whereas baseline severity was more dominant for Prostate Cancer. This indicates that even within a single broad disease category, depression outcome predictors may differ substantially.

Additional hierarchical analyses showed similarly distinct patterns in other clinical subtypes. For Predialysis patients, hospital-center factors were influential, while for patients with lower-limb amputation without prosthesis, therapy adherence was a prominent predictor. Some rare subgroups could not be modeled reliably because of data sparsity, highlighting the need for larger cohorts to support more granular subgroup inference.

\section{Conclusion}
The central finding of this work is that depression outcomes following mindfulness-based intervention are not explained by baseline severity alone. Although baseline BDI-II remained the most consistent predictor across both prediction horizons, the relative importance of other factors changed over time and across clinical groups. At 12 weeks, depression severity was more strongly associated with clinical condition and hospital-center context, whereas 24-week outcomes showed greater dependence on therapy engagement and demographic factors. This temporal shift suggests that short-term response may be shaped by immediate clinical and care-setting factors, while longer-term trajectories may depend more on sustained behavioral participation.

The predictive and interpretability results support this time-dependent and context-dependent interpretation. The best-performing 12-week model was an interpretable regularized linear model, whereas the best-performing 24-week model was a non-linear gradient-boosting model, suggesting that longer-term depression trajectories may involve more complex relationships among baseline severity, adherence, demographics, and clinical context. Disease-specific and hierarchical subgroup analyses further showed that predictor profiles differed across broad disease groups and within disease categories, particularly across cancer subtypes. These findings argue against a uniform intervention strategy and support personalized, context-aware mental-health pathways: adherence-dominant subgroups may benefit from engagement support, baseline-severity-dominant subgroups may require early psychological triage, and hospital-context-dominant subgroups may warrant review of institutional care practices.

Several limitations should be noted. The cohort size was modest, some granular clinical subgroups were too small for reliable modeling, and missing follow-up outcomes required model-based stochastic imputation before final train-test evaluation; therefore, predictive performance should be interpreted as exploratory rather than as strictly leakage-isolated validation. Future work should validate these findings in larger external cohorts, use train-only or nested imputation protocols, incorporate repeated-measure or mixed-effects models, and include additional variables such as medication, socioeconomic status, comorbidity burden, and social support. Despite these limitations, the study demonstrates that interpretable, subgroup-aware modeling can reveal clinically useful temporal and disease-specific patterns in depression outcome prediction.

\bibliographystyle{IEEEtran}
\bibliography{references}

\end{document}